\documentclass[accepted]{uai2021} % after acceptance, for a revised
\usepackage[american]{babel}
\usepackage{natbib} % has a nice set of citation styles and commands
\renewcommand{\bibsection}{\subsubsection*{References}}
\usepackage{mathtools} % amsmath with fixes and additions
\usepackage{booktabs} % commands to create good-looking tables
\usepackage{tikz} % nice language for creating drawings and diagrams
\usepackage{amsmath,amssymb,amsthm,textcomp}
\usepackage{multicol}
\usepackage{comment}
\usepackage[ruled,vlined]{algorithm2e}
\usepackage{algorithmic}
\usepackage{wrapfig}
\usepackage{subfig}
\usepackage{stackengine}
\usepackage{amssymb}% http://ctan.org/pkg/amssymb
\usepackage{pifont}% http://ctan.org/pkg/pifont
\usepackage{amsmath}
\DeclareMathOperator*{\argmax}{arg\,max}

\DeclareMathOperator*{\R}{\mathbb{R}}
\DeclareMathOperator*{\softmax}{softmax}

\newcommand{\norm}[1]{\left\lVert#1\right\rVert}

\title{Investigating Vulnerabilities of Deep Neural Policies}

\author[1]{\href{mailto:Ezgi Korkmaz <ezgikorkmazk@gmail.com>?Subject=Your UAI 2021 paper}{Ezgi Korkmaz}{}}
\affil[1]{%
    KTH Royal Institute of Technology \\
    Stockholm, Sweden.
}
\begin{document}
\maketitle

\begin{abstract}
Reinforcement learning policies based on deep neural networks are vulnerable to imperceptible adversarial perturbations to their inputs, in much the same way as neural network image classifiers. Recent work has proposed several methods to improve the robustness of deep reinforcement learning agents to adversarial perturbations based on training in the presence of these imperceptible perturbations (i.e. adversarial training). In this paper, we study the effects of adversarial training on the neural policy learned by the agent. In particular, we follow two distinct parallel approaches to investigate the outcomes of adversarial training on deep neural policies based on worst-case distributional shift and feature sensitivity. For the first approach, we compare the Fourier spectrum of minimal perturbations computed for both adversarially trained and vanilla trained neural policies. Via experiments in the OpenAI Atari environments we show that minimal perturbations computed for adversarially trained policies are more focused on lower frequencies in the Fourier domain, indicating a higher sensitivity of these policies to low frequency perturbations. For the second approach, we propose a novel method to measure the feature sensitivities of deep neural policies and we compare these feature sensitivity differences in state-of-the-art adversarially trained deep neural policies and vanilla trained deep neural policies. We believe our results can be an initial step towards understanding the relationship between adversarial training and different notions of robustness for neural policies.

\end{abstract}
\section{Introduction}\label{sec:intro}

Deep neural networks (DNNs) have led to notable progress across many areas of machine learning research and applications including computer vision \cite{alex12}, natural language processing \cite{ilya14}, and speech recognition \cite{hannun14}. More recently deep neural networks (DNNs) have been employed in deep reinforcement learning by \citet{mn15} to approximate the state-action value function for large action size or state size MDPs. With this initial success deep reinforcement learning became an emerging subfield with many applications such as robotics \citet{levine18}, financial trading \citet{noonan17} and medical \citet{dao18, tseng17}.

While the success of DNNs grew, a line of research focused on their reliability and robustness. Initially, \citet{szegedy13} demonstrated that it is possible to fool image classifiers by adding visually imperceptible perturbations to neural network inputs. Follow up work by \citet{fellow15} showed that these perturbations demonstrate that deep neural networks are learning linear functions. Several studies focused on overcoming this susceptibility towards specifically crafted visually imperceptible perturbations, and proposed training neural networks to be robust to these worst-case perturbations \citet{madry18}.
However, adversarial training also has drawbacks: adversarially trained models tend to have lower accuracy on standard inputs, learn inaccurate suboptimal policies, and may be less robust to other types of distribution shift beyond worst-case $\ell_p$-norm bounded perturbations \citet{hong19, korkmaz21gen, korkmaz21cvpr}.
While there is a significant amount of study focusing on adversarial training several works suggest that the existence of adversarial perturbations may be inevitable \cite{elvis19, saed19,pascal19}.

In this paper we focus on searching for answers to the following questions: (i) What are the sensitivity differences between state-of-the-art adversarially trained deep neural policies and vanilla trained deep neural policies at a high level? (ii) Does adversarial training move the sensitivities to worst-case $\ell_p$-norm bounded distributional shift towards different directions in the input for deep neural policies? (iii) Does adversarial training create a new set of non-robust features while eliminating the existing ones? Therefore, in this paper we focus on examining the affects of adversarial training on neural policies in deep reinforcement learning and make the following contributions:

\begin{itemize}
\item We investigate the properties of adversarially trained neural policies via two different perspectives. The first is based on the response of adversarially trained policies to worst-case perturbations, and the second is based on probing adversarially trained policies via their sensitivity to features.
\item For the worst-case perspective, we compare the frequency domain of the perturbations produced by \citet{carlini17} for adversarially trained models and vanilla trained models.
\item We show that the perturbations produced from adversarially trained models are suppressed in high frequencies and more concentrated in lower frequencies in the Fourier domain compared to vanilla trained neural policies.
\item For the sensitivity perspective, we propose a novel algorithm to detect feature based vulnerabilities of trained deep neural policies.
\item We show that adversarially trained policies have a distinctive sensitivity pattern compared to vanilla trained deep neural policies. Furthermore, we demonstrate that while adversarially training removes the sensitivity of the neural policies towards some non-robust features, it also creates sensitivity towards a new set of non-robust features.
\end{itemize}

\section{Background}

\subsection{Preliminaries}
In this paper we focus on deep reinforcement learning for Markov decision processes (MDPs) given by a set of continuous states $S$, a set of discrete actions $A$, a transition probability distribution $P$ on $S\times A \times S$, and a reward function $r:S \times A \to \R$. A policy $\pi:S\to \mathcal{P}(A)$ for an MDP assigns a probability distribution on actions to each $s \in S$. The goal for the reinforcement learning agent is to learn a policy $\pi$ that maximizes the expected cumulative discounted reward $R = \mathbb{E}\sum_{t=0}^{T-1}\gamma^tr(s_t,a_t)$ where $a_t \sim \pi(s_t)$. In $Q$-learning the learned policy is parametrized by a state-action value function $Q:S\times A\to \R$, which represents the value of taking action $a$ in state $s$. Let $a^*(s) = \argmax_a Q(s,a)$ denote the highest $Q$-value for an action in state $s$. The $\epsilon$-greedy policy of the agent for $Q$-learning is given by taking action $a^*(s)$ with probability $1-\epsilon$, and a uniformly random action with probability $\epsilon$.

\subsection{Adversarial Examples}

Manipulating the output of neural networks by adding imperceptible perturbations was introduced by \citet{szegedy13} based on a box constrained optimization method. While this proposed method was computationally expensive, \citet{fellow15} proposed a faster and simpler method based on gradients in a nearby $\epsilon$-ball,

\begin{equation}
\mathnormal{\displaystyle x_{\textrm{adv}} = x+ \epsilon \cdot \frac{\nabla_{x}J(\displaystyle x,y)}{||\nabla_{x}J(x,y)||_p},}
\end{equation}

where $x$ represents the input, $y$ represents the labels, and $J(x,y)$ represents the cost function used to train the network. \citet{kurakin16} further proposed an iterative search method inside this $\epsilon$-ball using the fast gradient sign method (FGSM) proposed by \citet{fellow15}.

\begin{gather}
x^0_{\textrm{adv}}= x,\\
x_{\textrm{adv}}^{N+1} = \textrm{clip}_\epsilon(x_{\textrm{adv}}^N +\alpha \textrm{sign}(\nabla_x J(x^N_{\textrm{adv}},y)))
\end{gather}

This method is also known as projected gradient descent (PGD) as in \citet{madry18}. \citet{carlini17} formulated the problem of producing adversarial perturbations in a more targeted way and proposed a method based on distance minimization for a given label in image classification. For deep reinforcement learning this formulation is based on distance minimization for a given a target action which is not equal to the best action decided by the trained policy,

\begin{equation*}
\begin{aligned}
&& \underset{s_{\textrm{adv}} \in D_{\epsilon,p}(s)}{\text{min}}
&  \mathnormal {\|s_{\textrm{adv}}-s\|_p}\\
& \text{subject to}
& \argmax_{a} & Q(s,a) \neq \argmax_{a}Q(s_{\textrm{adv}},a),
\end{aligned}
\label{lin}
\end{equation*}

Note that $Q(s,a)$ denotes the state-action value function of the deep neural policy. \citet{athalye} showed that the \citet{carlini17} adversarial formulation can break several proposed defenses. For this reason, in this paper we will focus on perturbations produced by the \citet{carlini17} formulation.

\subsection{Perturbation Formulations and Adversarial Training}

Initially adversarial examples were introduced in the deep reinforcement learning domain by \citet{huang17} and \citet{kos17} concurrently by utilizing FGSM as
\begin{figure*}[t]
\footnotesize
\begin{center}
\stackunder[6pt]{\includegraphics[scale=0.245]{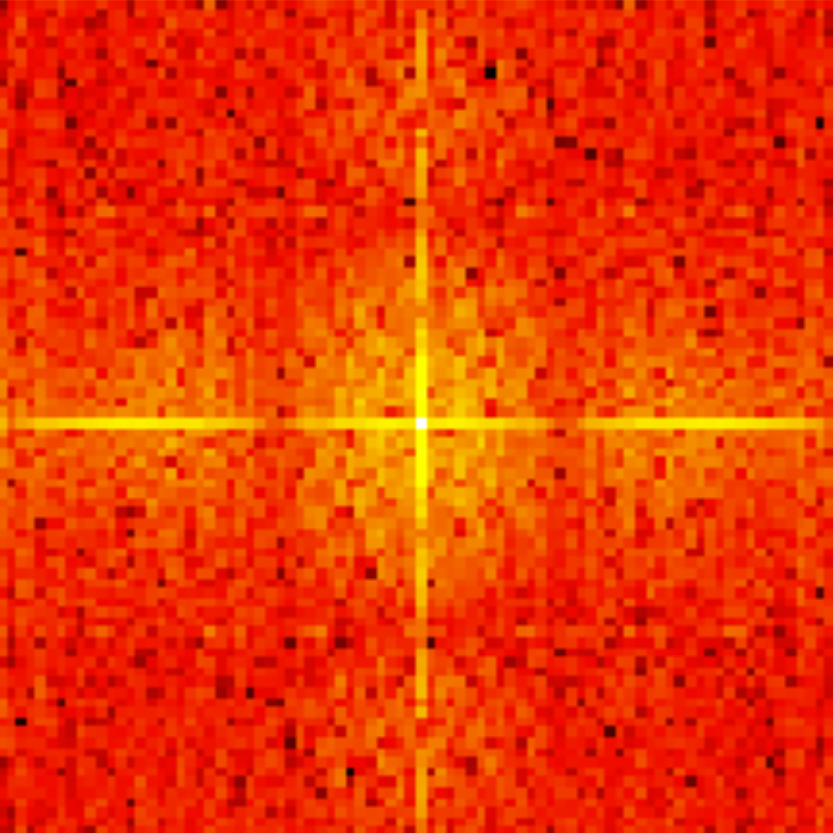}}{Freeway}
\hskip 0.1pt
\stackunder[6pt]{\includegraphics[scale=0.225]{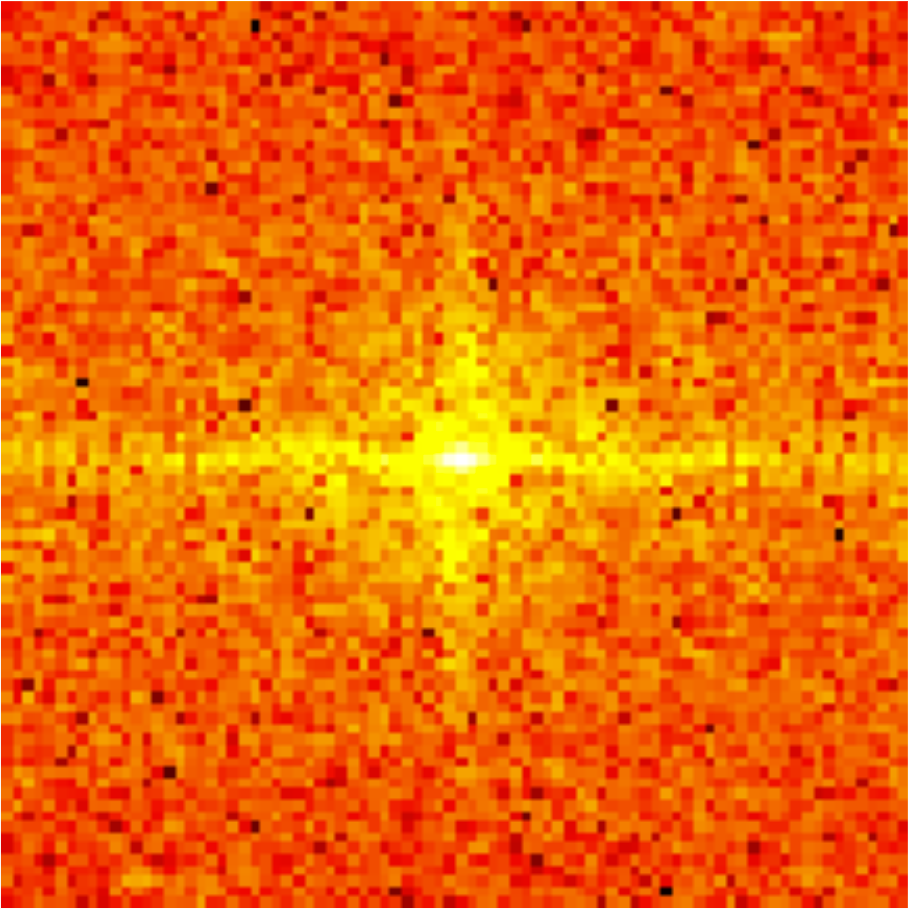}}{RoadRunner}
\stackunder[6pt]{\includegraphics[scale=0.215]{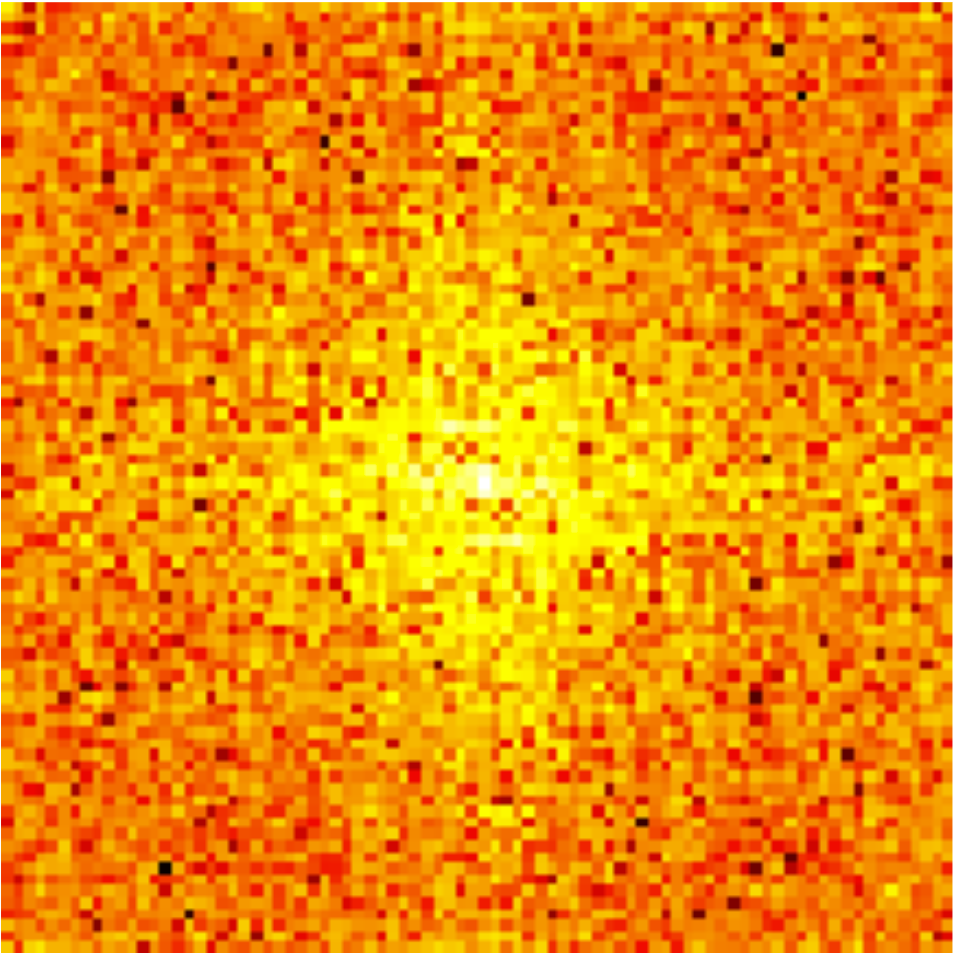}}{BankHeist}\\
\stackunder[6pt]{\includegraphics[scale=0.24]{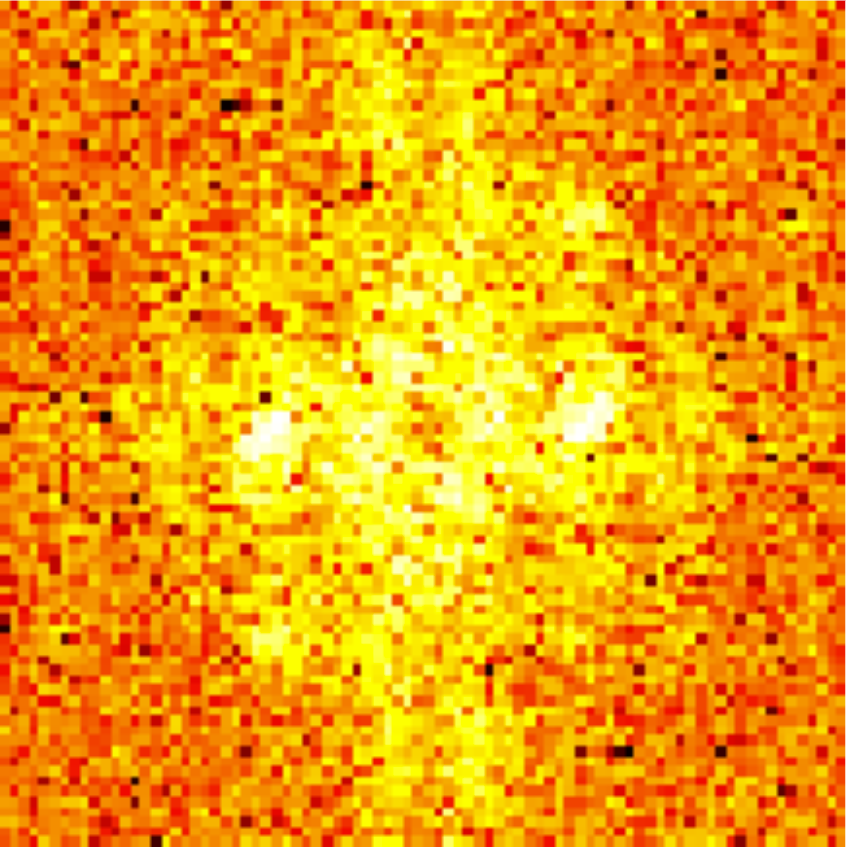}}{Freeway}
\stackunder[6pt]{\includegraphics[scale=0.22]{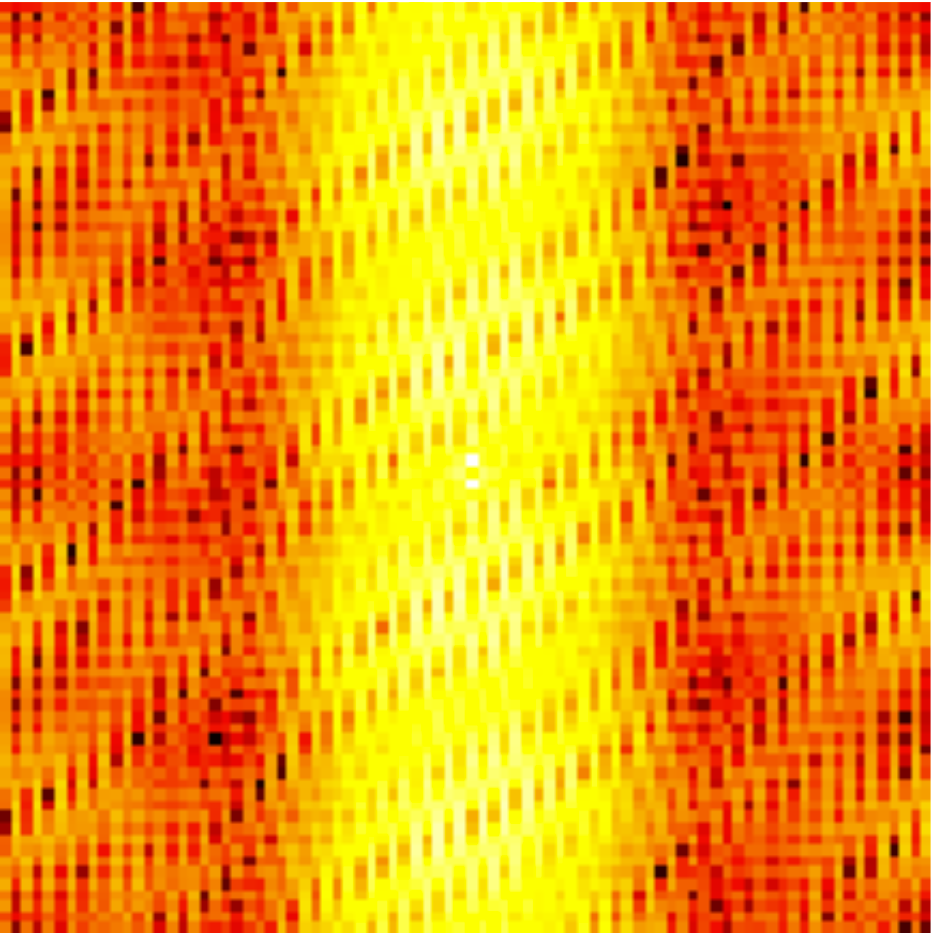}}{RoadRunner}
\stackunder[6pt]{\includegraphics[scale=0.22]{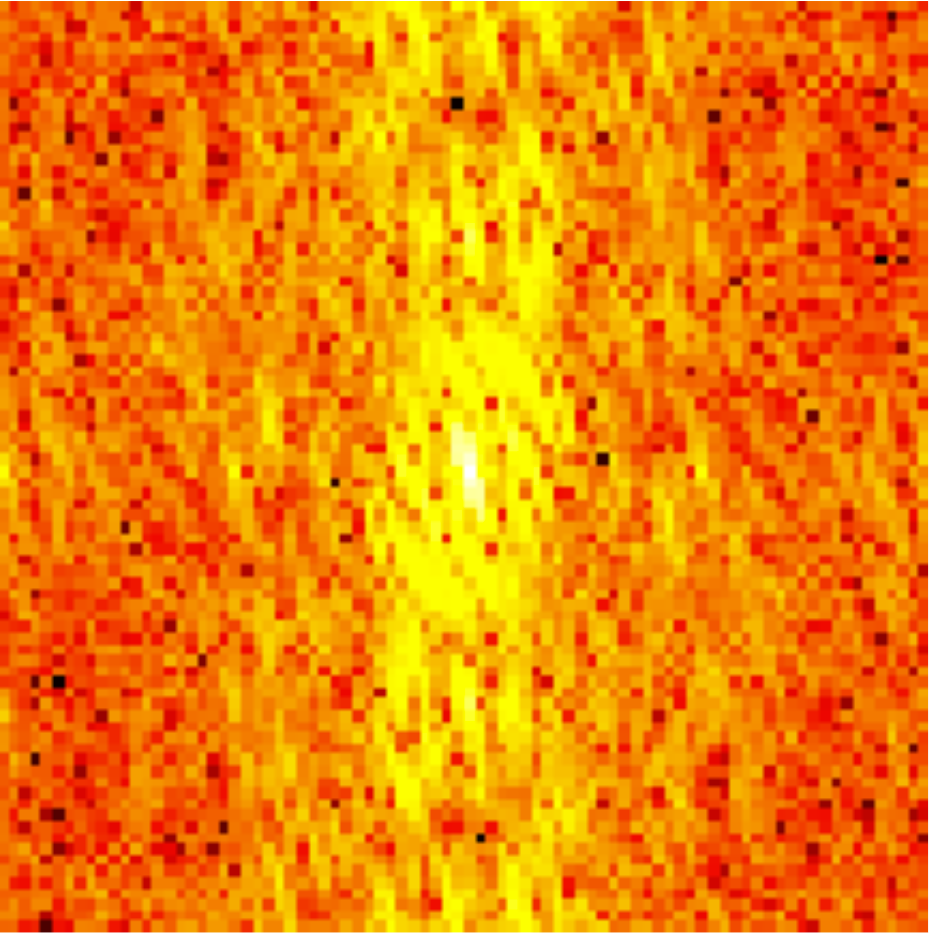}}{BankHeist}
\end{center}
  \caption{Fourier spectrum of the perturbations computed via \citet{carlini17} for state-of-the-art adversarially trained models and vanilla trained models. First Row: Adversarially trained. Second Row: Vanilla trained.}
    \label{perturbations}
\end{figure*}

proposed by \citet{fellow15}. Several studies have been conducted to make deep reinforcement learning policies more robust to such specifically crafted malicious examples. \citet{mandlekar17} use adversarial examples produced by FGSM in the training data  to regularize the policy in an attempt to increase robustness. \citet{pinto17} model the interaction between the perturbation maker and the agent as zero-sum Markov game and proposes a joint training algorithm to improve robustness against an adversary which aims to minimize the expected cumulative reward of the agent. \citet{glaeve19} model the relationship between the adversary and the agent as zero-sum game where the adversary is limited to taking natural actions in the environment rather than minimal $\ell_p$-norm bounded perturbations, and proposes an approach based on self-play to gain robustness against such an adversary. Quite recently \citet{huan20} proposed a modified MDP called state-adversarial MDP with the aim of obtaining theoretically principled robust polices towards natural measurement errors and $\ell_p$-norm worst case perturbations.

\section{Investigating Vulnerabilities}

In this paper we aim to seek answers for several questions:

\begin{itemize}
\item What are the susceptibility differences between state-of-the-art adversarially trained deep neural policies and vanilla trained deep neural policies?
\item Do the sensitivities of deep neural policies shifts from worst-case $\ell_p$-norm bounded perturbations towards different directions in the input with adversarial training?
\item Does adversarial training create a new set of non-robust features while eliminating the existing ones?
\end{itemize}

In our experiments we use OpenAI \citet{openai} Atari baselines \citet{bell13}. Our models are trained with DDQN \citet{wang16} and SA-DDQN \citet{huan20}. We test trained policies averaged over 10 episodes. Note that SA-DDQN is certified against $\ell_\infty$-norm bounded at $1/255$. Therefore, we also bound the perturbation by this threshold and find the perturbations with $\ell_\infty$-norm lower than this value.

\section{Neural Policy Perturbations in the Fourier Domain}

For the adversarially trained agents, we focus on the state-of-the-art adversarial training algorithm SA-DQN proposed by \citet{huan20}. In this study the authors model the interaction between the neural policy and the introduced perturbations as a state-adversarial modified Markov Decision Process (MDP). The authors claim that the agents trained in SA-MDP with the proposed algorithm SA-DQN are more robust to adversarial perturbations and natural noise introduced to the agent's perception system. Furthermore, the authors demonstrate the robustness of SA-DQN against perturbations produced by the PGD attack proposed by \citet{madry18}.

\begin{figure*}[h!]
\begin{multicols}{3}
\includegraphics[width=1.1\linewidth]{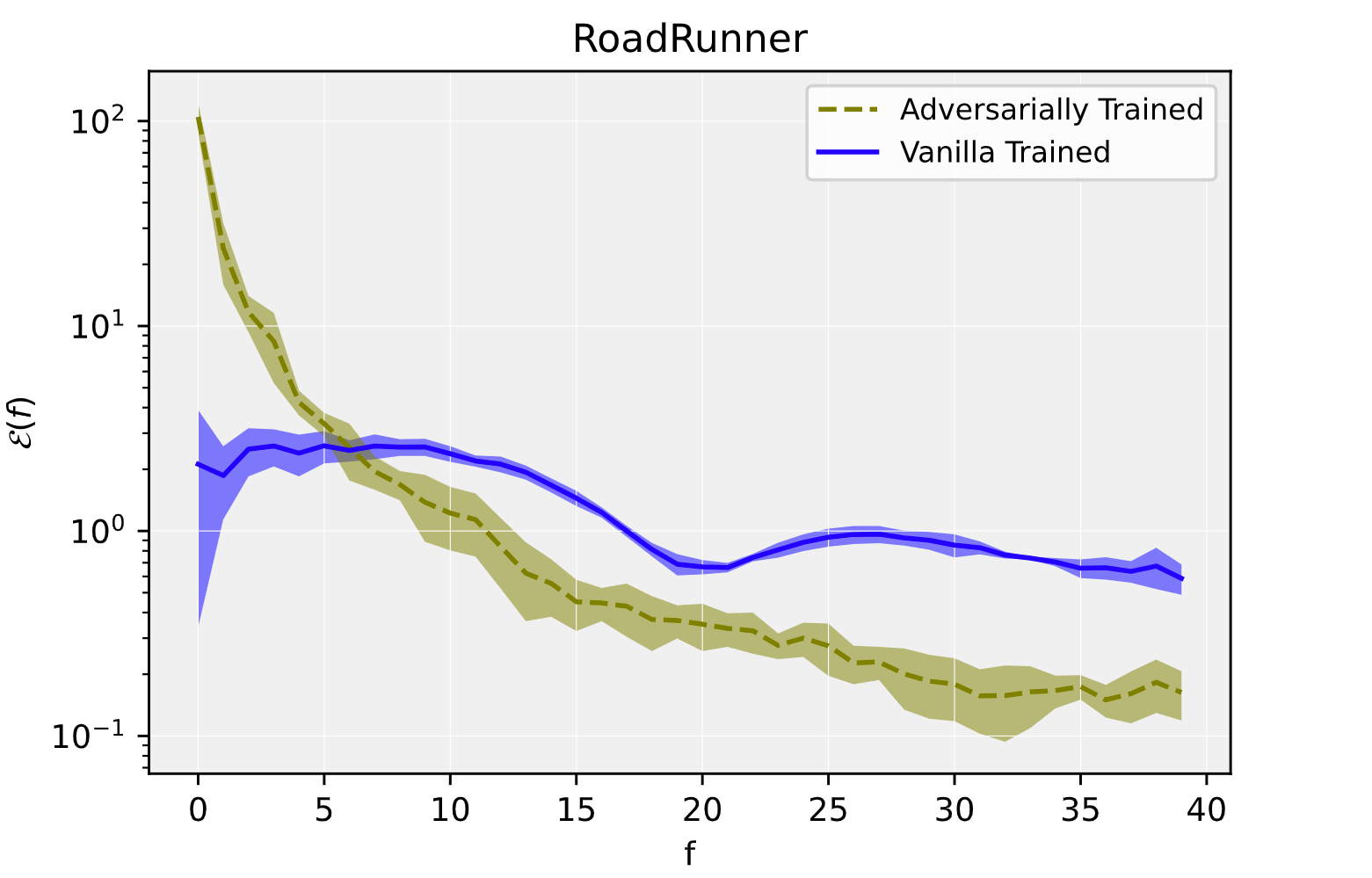}\par
\includegraphics[width=1.1\linewidth]{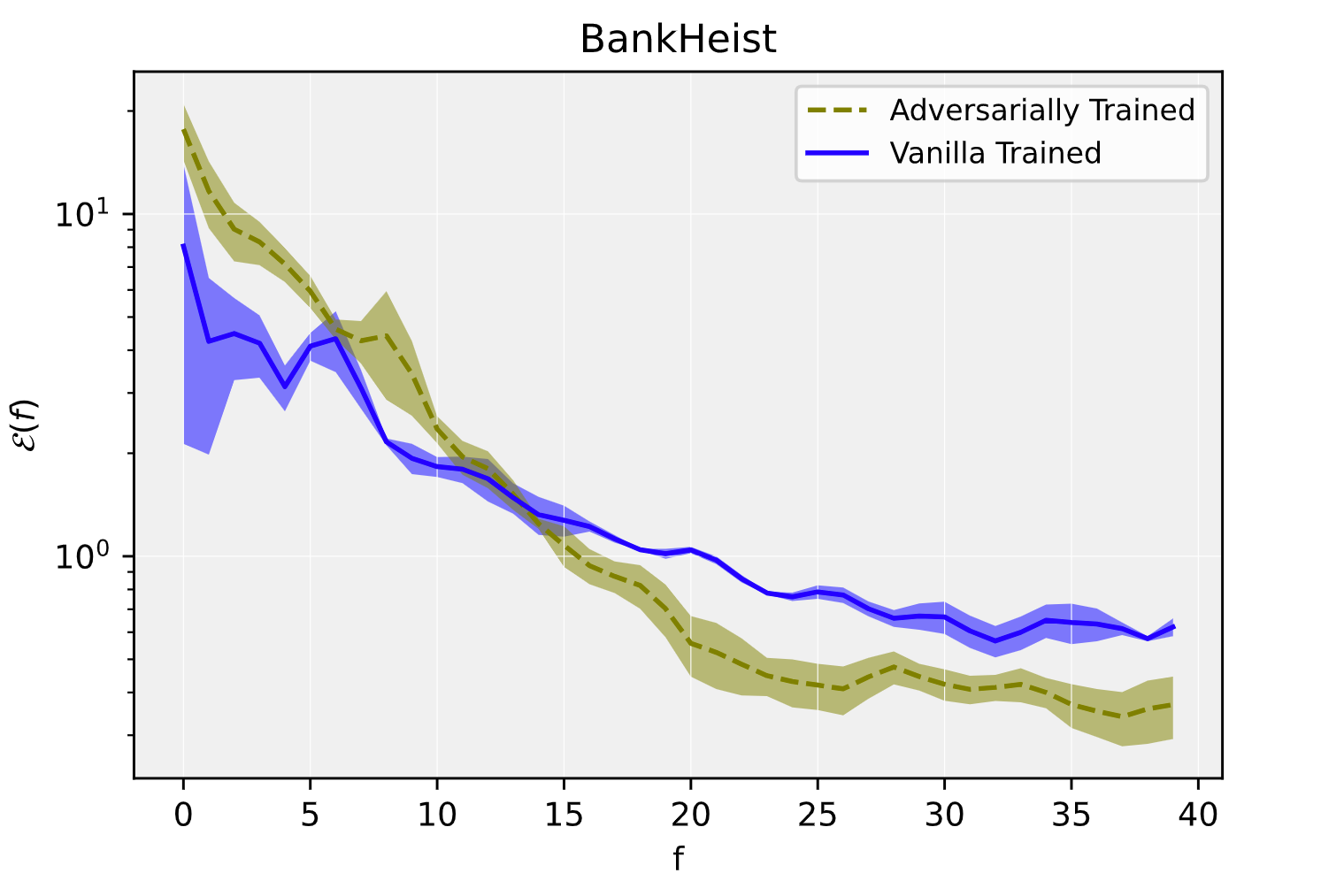}\par
\includegraphics[width=1.1\linewidth]{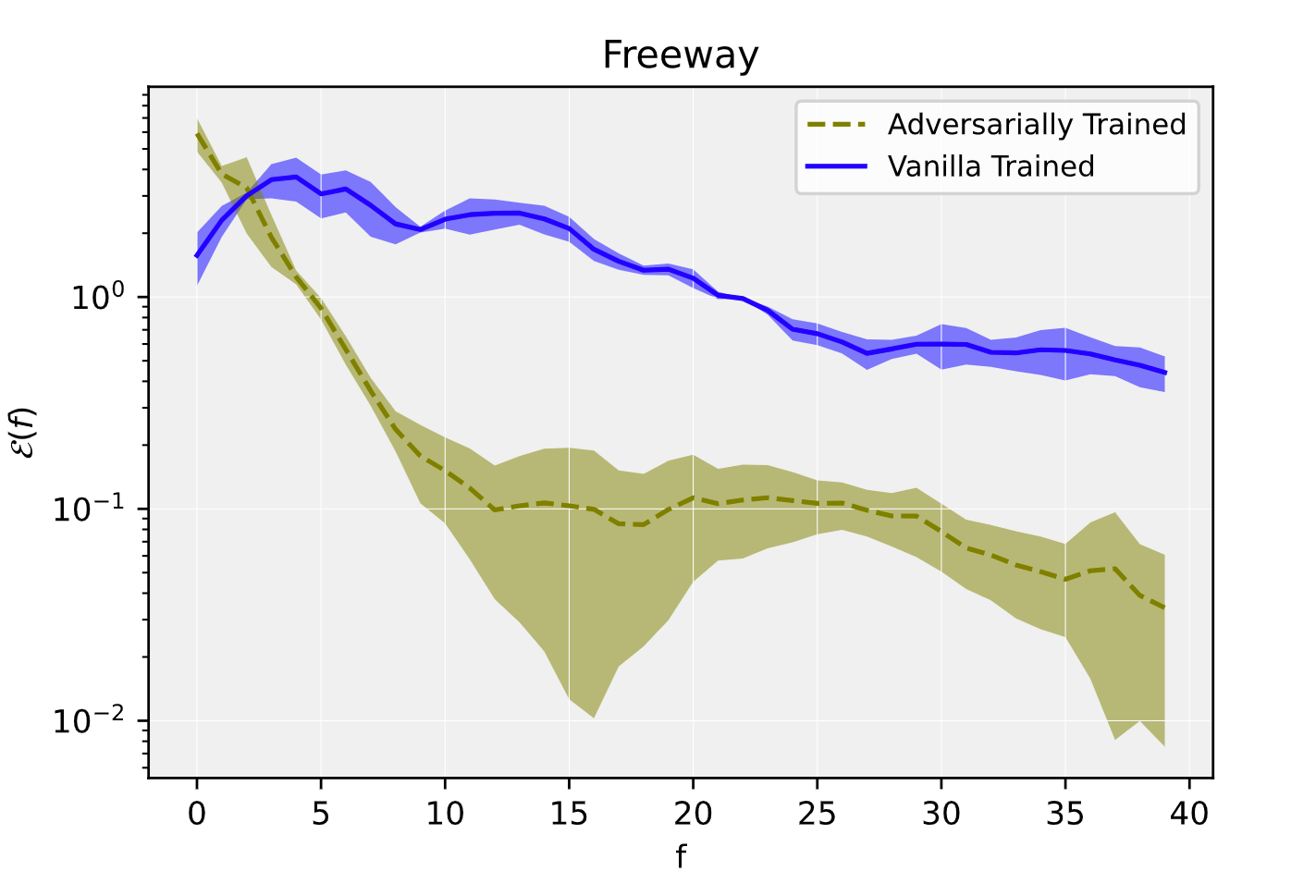}\par
\end{multicols}
\vspace{-0.4cm}
\caption{Power spectrum of the perturbations computed via \citet{carlini17} for adversarially trained models and vanilla trained models in Fourier domain for RoadRunner, BankHeist and Freeway.}
\label{advtrainrotation}
\end{figure*}

In this section we conduct an investigation on the frequency domain of the perturbations computed from vanilla trained agents and adversarially trained agents.
In particular, we compute a minimum length perturbation via \citet{carlini17} which causes the agent to change its optimal learned action. We found that the certified defence proposed by \citet{huan20} can be overcome via \citet{carlini17} for the certified bound given in \citet{huan20}. For these minimal perturbations we compute the Fourier transform of the perturbation and record this data. By comparing the results of this experiment for adversarially trained versus vanilla trained agents, we can understand the affects of adversarial training on the directions to which the neural policy is sensitive. We now describe the details of the experimental setup.

In more detail, we ensure that the perturbations $\eta = s_{\rm{adv}} - s$ produced by \citet{carlini17} satisfies two requirements:
\begin{itemize}
\item The optimal action in state $s$ changes i.e. $a^*(s) \neq a^*(s_{\rm{adv}})$.
\item The perturbation is bounded by the certified defense level proposed by \citet{huan20} $\lVert\eta\rVert_\infty < \frac{1}{255}$.
\end{itemize}

In Figure \ref{perturbations} we visualize the Fourier transform of a minimal perturbation for both vanilla trained and adversarially trained agents in RoadRunner, BankHeist, and Freeway. The center of each image corresponds to the Fourier basis function where both spatial frequencies are zero, and the magnitude of the spatial frequencies increases as one moves outward from the center. There is a distinctive difference between vanilla and adversarially trained agents in the qualitative appearance of Fourier transforms of the minimal perturbations. Further, from these visualizations it is clear that the perturbations for the adversarially trained models generally have their Fourier transform concentrated at lower frequencies than those of the vanilla trained models.

To make this claim formal, for each number $f$ we compute the total energy $\mathcal{E}(f)$ of the Fourier transform for basis functions whose maximum spatial frequency is equal to $f$. In Figure \ref{advtrainrotation} we plot the average of $\mathcal{E}(f)$ over the minimal perturbations computed in our experiments.
 We find that the minimal perturbations computed for adversarially trained neural policies are indeed shifted towards lower frequencies when compared to those for vanilla trained neural policies. This shift in the frequency domain of the computed perturbations implies that adversarially trained neural policies may be more sensitive towards low frequency perturbations.

\section{Visualizing Neural Policy Vulnerabilities}
\label{fourier}

In this section we propose two different methods to visualize vulnerabilities of deep neural policies to their input observations. First, we describe our proposed method of feature vulnerability mapping KMAP in detail. To be able to visualize weaknesses we record the drop in the state-action value $Q(s,a)$ caused by setting each pixel in $s$ to zero one at a time. In particular, let $Z_{i,j}:S \to S$ be the function which sets the $i,j$ coordinate of $s$ to zero and leaves the other coordinates unchanged. We define,

\begin{equation}
\label{k}
\mathcal{K}(i,j) = Q(s,a^*) - Q(s,\argmax_{a} Q(Z_{i,j}(s),a)).
\end{equation}

Note that the difference in Equation \ref{k} represents the drop in the $Q$-value in state $s$, when taking the optimal action for the state $Z_{i,j}(s)$. Therefore, $\mathcal{K}(i,j)$ aims to measure the drop in the $Q$-values of the neural policy with respect to individual pixel changes. In other words, $\mathcal{K}(i,j)$ is a mapping of features to an importance metric determined by the deep neural policy. We describe our proposed KMAP method in detail in Algorithm \ref{kmap}.

As a natural point of comparison we propose another algorithm HMAP to visualize input based vulnerabilities. In particular, HMAP is based on measuring the effect of each individual pixel on the decision of the deep neural policy by measuring the cross-entropy loss between $\pi(s,a)$ and $\pi(Z_{i,j}(s),a)$.

\begin{algorithm}[h]
   \caption{KMAP Feature vulnerability mapping}
   \label{kmap}
\SetAlgoLined
\begin{algorithmic}
   \STATE {\bfseries Input:} State-action value function $Q(s,a)$,actions $a$, states $s$,  $T_d$ size of the dimension $d$ of the state $s$, and $s(i,j)$ is the value of the $i,j$-th pixel in state $s$.
    \STATE {\bfseries Output:} Visual weakness mapping function $\mathcal{K}(i,j)$
    \STATE $s_{\textrm{aug}} = s$
   \FOR{$i=1$ {\bfseries to} $T_1$}
   \FOR{$j=1$ {\bfseries to} $T_2$}
    \STATE $s_{\textrm{aug}}(i,j) = 0$
   \STATE $a_{\textrm{aug}}^*$ = argmax$_{a}$ $Q(s_{\textrm{aug}},a)$
   \STATE $a^*$ = argmax$_{a}$ $Q(s,a)$
   \STATE $\mathcal{K}(i,j)$ += $Q(s,a) - Q(s,a_{\textrm{aug}}^*)$
   \STATE $s_{\textrm{aug}} = s$
   \ENDFOR
   \ENDFOR
   \STATE {\bfseries Return:} $\mathcal{K}(i,j)$
\end{algorithmic}
\end{algorithm}

\begin{equation}
\label{h}
\mathcal{H}(i,j) = -\sum_{a \in A} \pi(s,a) \log(\pi(Z_{i,j}(s),a))
\end{equation}
where we compute the policy $\pi(s,a)$ via the softmax of the state-action value function $Q(s,a)$,

\begin{equation}
\label{pi}
\pi(s,a) = \dfrac{e^{Q(s,a)/T}}{\sum_{a \in A}e^{Q(s,a)/T}}.
\end{equation}

Note that $T$ represents the temperature constant. We describe the HMAP method in detail in Algorithm \ref{hmap}.

\section{Results on KMAP and HMAP}

\begin{figure*}[h!]
\footnotesize
\begin{center}
\stackunder[6pt]{\includegraphics[scale=0.33]{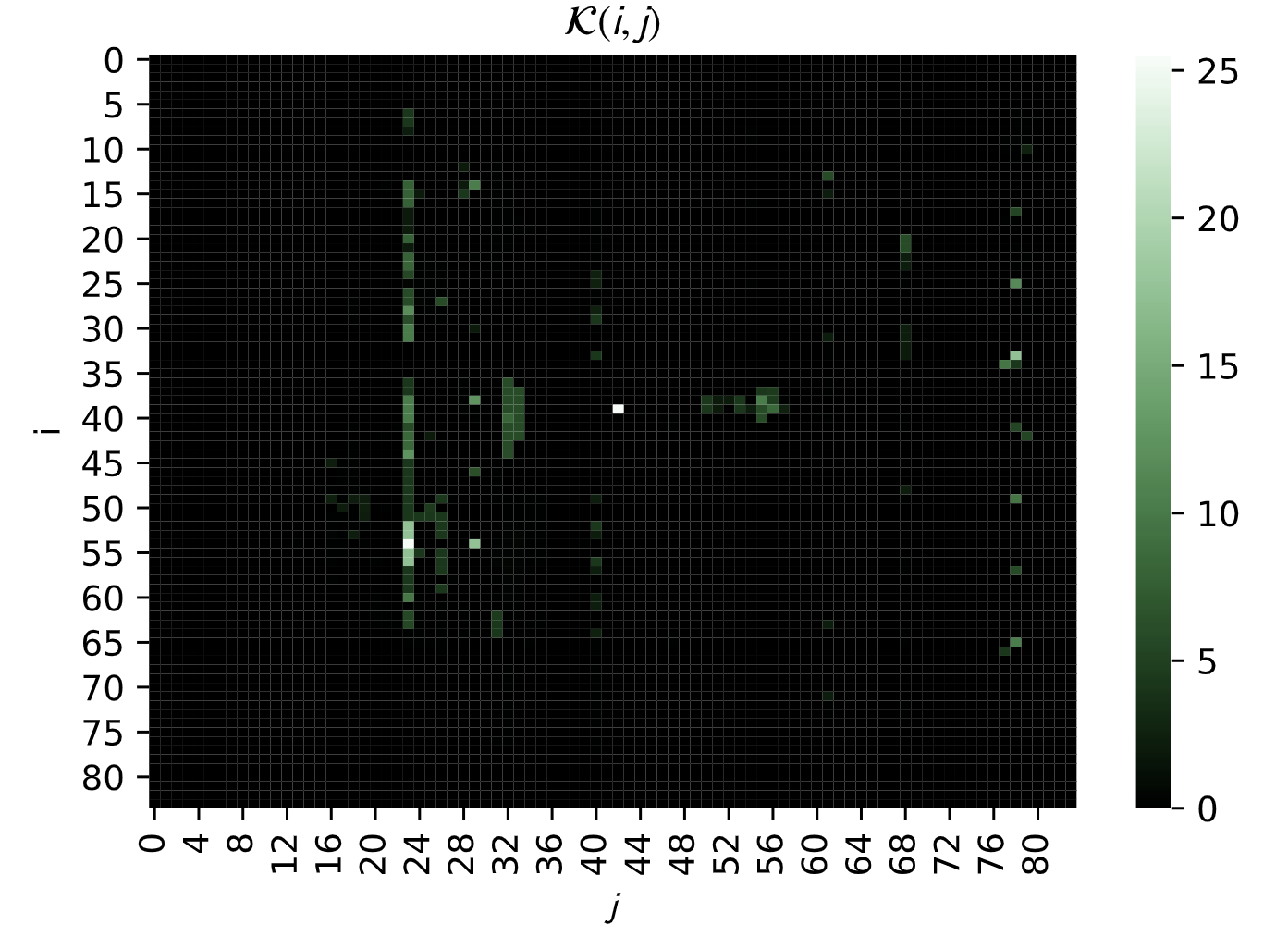}}{Adversarially Trained}
\hskip 0.1pt
\stackunder[6pt]{\includegraphics[scale=0.33]{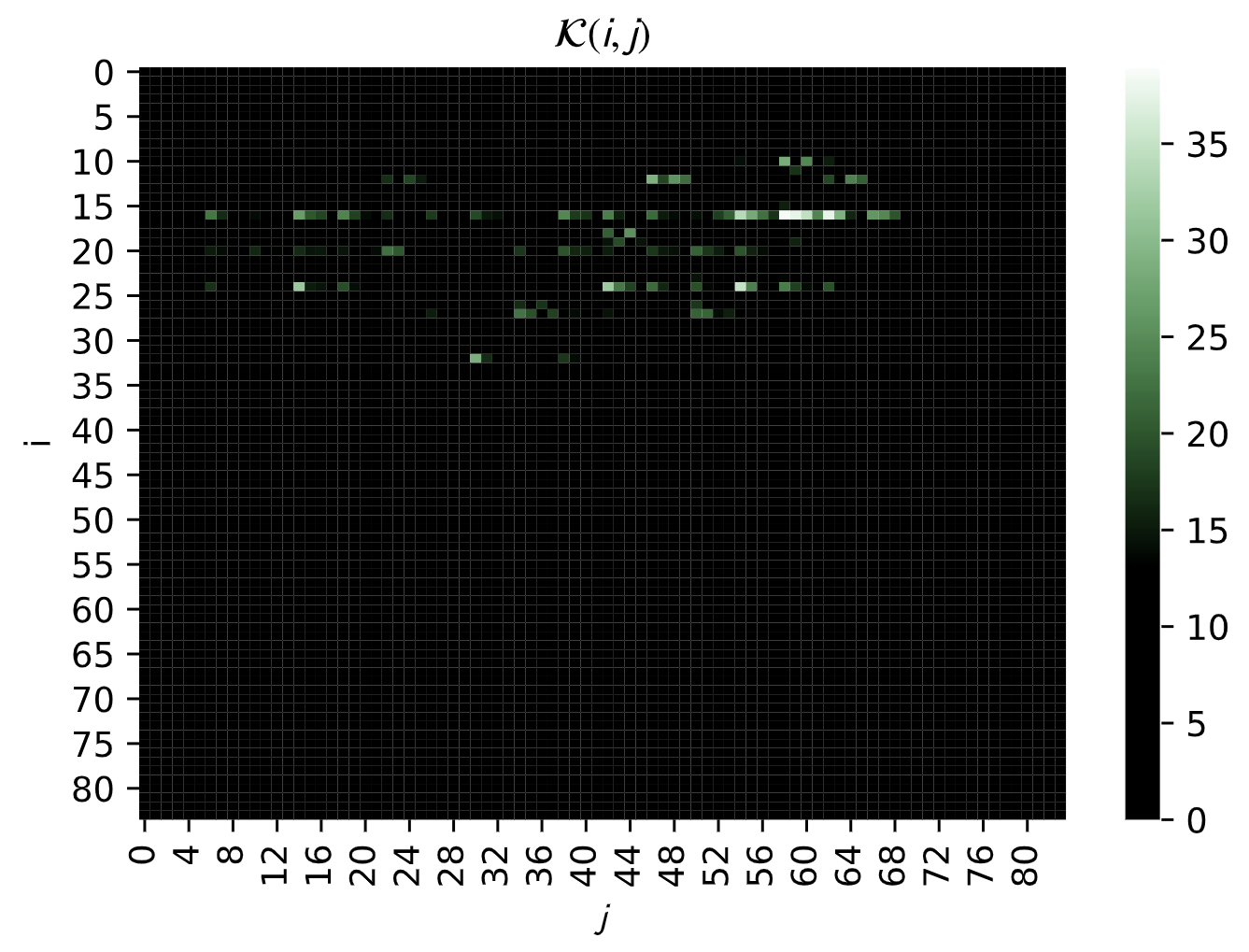}}{Vanilla Trained}
\end{center}
  \caption{KMAP $\mathcal{K}(i,j)$ heatmaps for state-of-the-art adversarially trained deep neural policy and vanilla trained deep neural policy in RoadRunner. Left: Adversarially trained. Right: Vanilla trained.}
    \label{roadkmap}
\end{figure*}

In Figure \ref{roadkmap} we show the KMAP and HMAP heatmaps for a state-of-the-art adversarially trained neural policy and vanilla trained neural policy in RoadRunner. One intriguing observation from the KMAP heatmap for the adversarially trained deep neural policy is the vulnerability to pixel changes in a certain column shown in Figure \ref{roadkmap}. In comparison, the vanilla trained agent's vulnerability is concentrated on several rows in a different part of the input. Another interesting fact about Figure \ref{roadkmap} is that the vulnerability pattern for the vanilla trained agent is concentrated on a portion of the input image with which the agent does not interact during the game. In fact, in RoadRunner, the vulnerability pattern for the vanilla trained agent is in a portion of the input that the agent is not able to even visit.

\begin{figure*}[h!]
\footnotesize
\begin{center}
\stackunder[6pt]{\includegraphics[scale=0.34]{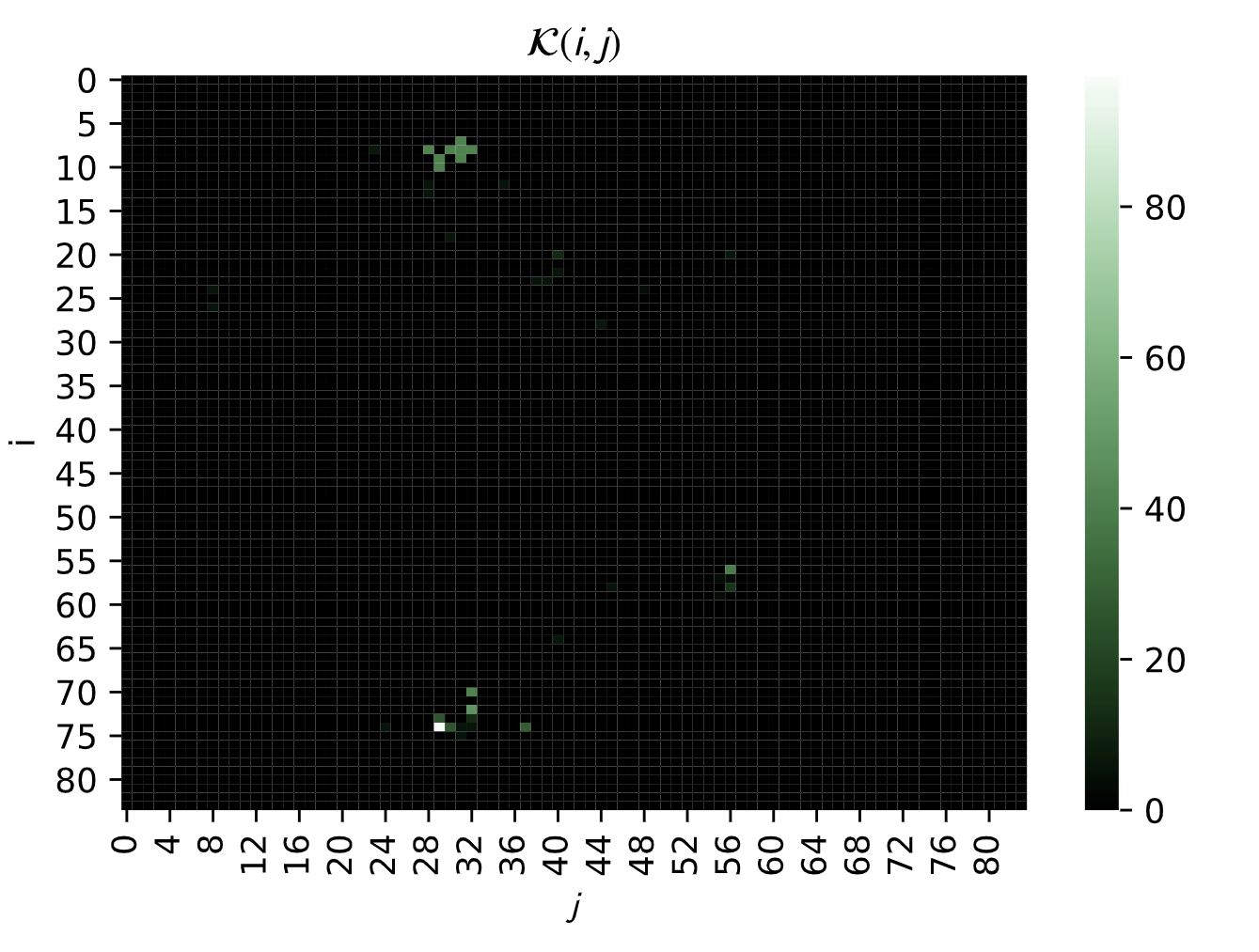}}{Adversarially Trained}
\hskip 0.1pt
\stackunder[6pt]{\includegraphics[scale=0.33]{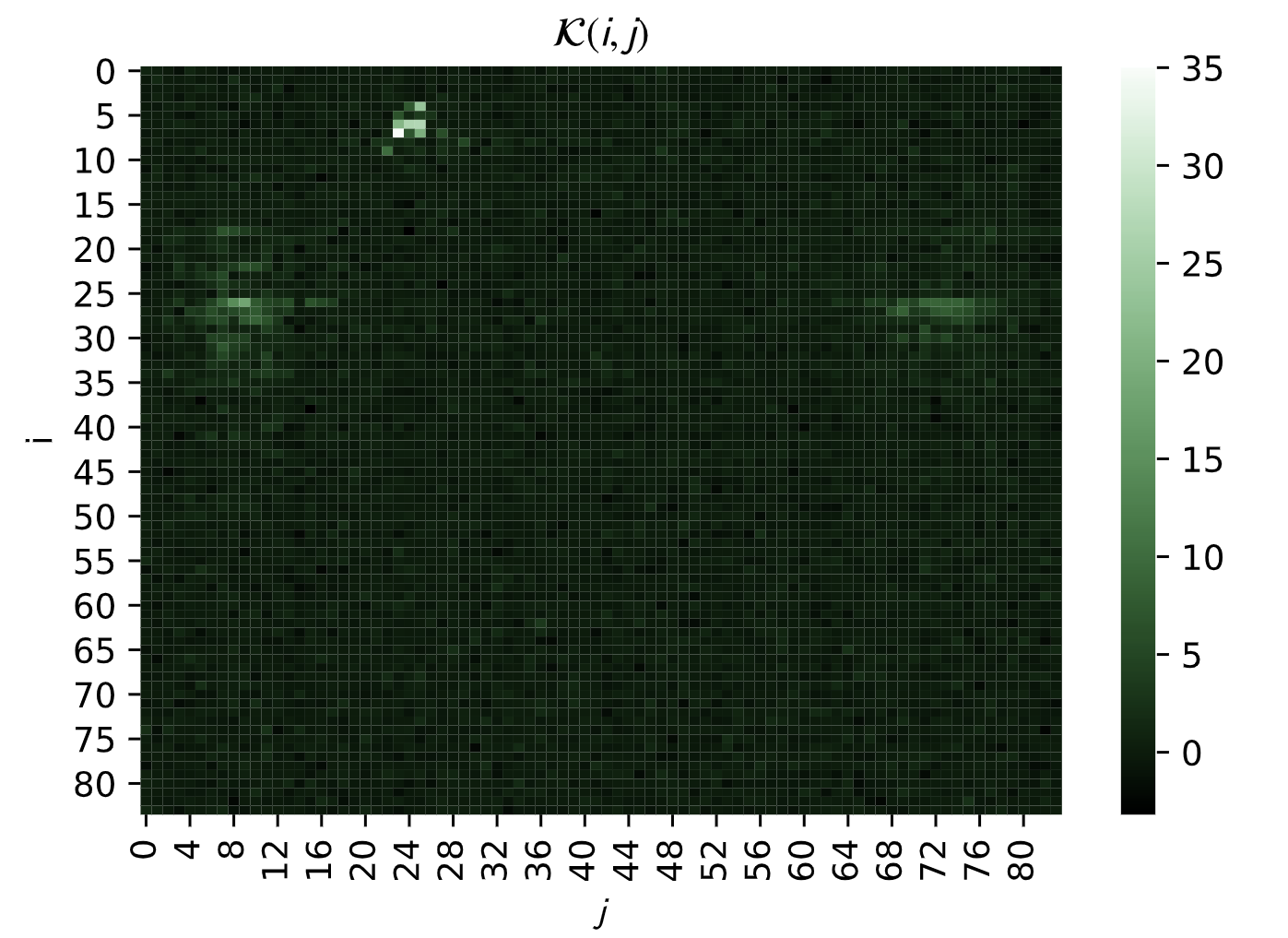}}{Vanilla Trained}
\end{center}
  \caption{KMAP $\mathcal{K}(i,j)$ heatmaps for state-of-the-art adversarially trained deep neural policy and vanilla trained deep neural policy in BankHeist. Left: Adversarially trained. Right: Vanilla trained.}
    \label{bankkmap}
\end{figure*}

Figure \ref{bankkmap} the BankHeist KMAP $\mathcal{K}(i,j)$ results show a similar sensitivity pattern between the adversarially trained deep neural policy and the vanilla trained deep neural policy. In particular adversarially trained KMAP $\mathcal{K}(6:10,29:31)$\footnote{$\mathcal{K}(k:m,l:n)$ denotes $\mathcal{K}(i,j)$ values for the set of coordinates $i \in \{k,\dots,m\}, j \in \{l,\dots,n\}$.} is quite similar to vanilla trained $\mathcal{K}(3:7,23:25)$\footnote{This portion of the input observation corresponds to the fuel gauge in BankHeist. In this game the player loses a life when the fuel runs out.}. Thus,

\begin{algorithm}[h!]
   \caption{HMAP Feature vulnerability mapping}
   \label{hmap}
\SetAlgoLined
\begin{algorithmic}
   \STATE {\bfseries Input:} State-action value function $Q(s,a)$, actions $a$, states $s$, policy $\pi(s,a)$, $T_d$ size of the dimension $d$ of the state $s$, and $s(i,j)$ is the value of the $i,j$-th pixel in state $s$.
    \STATE {\bfseries Output:} Visual weakness mapping function $\mathcal{H}(i,j)$
    \STATE $s_{\textrm{aug}} = s$
   \FOR{$i=1$ {\bfseries to} $T_1$}
   \FOR{$j=1$ {\bfseries to} $T_2$}
    \STATE $s_{\textrm{aug}}(i,j) = 0$
    \STATE $\pi(s,a) = \softmax(Q(s,a))$
    \STATE $\pi(s_{\textrm{aug}},a) = \softmax(Q(s_{\textrm{aug}},a))$
   \STATE $\mathcal{H}(i,j)$ +=$ -\sum_{a \in A} \pi(s,a) \log(\pi(s_{\textrm{aug}},a))$
   \STATE $s_{\textrm{aug}} = s$
   \ENDFOR
   \ENDFOR
   \STATE {\bfseries Return:} $\mathcal{H}(i,j)$
\end{algorithmic}
\end{algorithm}

\begin{figure*}[h!]
\footnotesize
\begin{center}
\stackunder[6pt]{\includegraphics[scale=0.33]{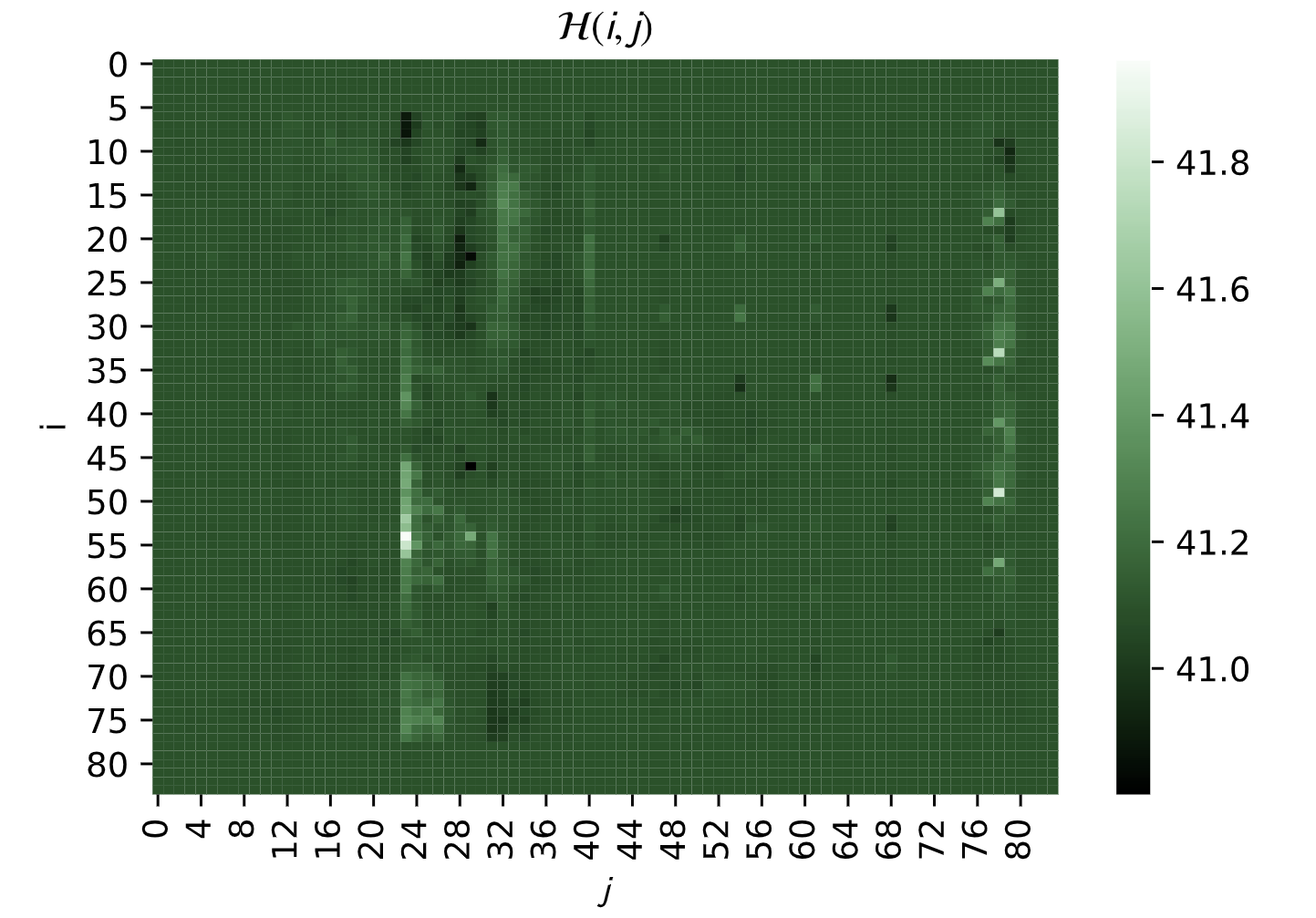}}{Adversarially Trained}
\hskip 0.1pt
\stackunder[6pt]{\includegraphics[scale=0.34]{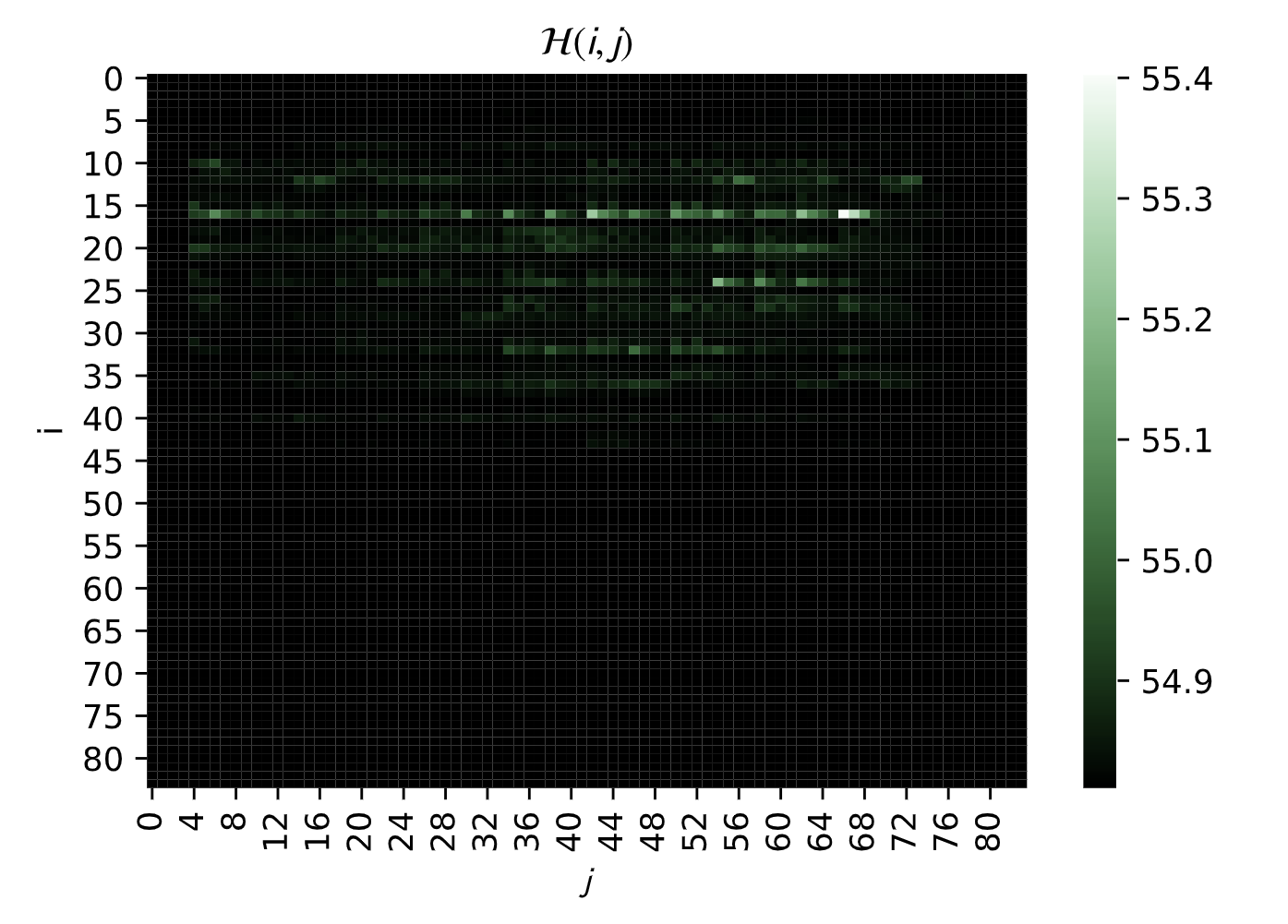}}{Vanilla Trained}
\end{center}
  \caption{HMAP $\mathcal{H}(i,j)$ heatmaps for state-of-the-art adversarially trained deep neural policy and vanilla trained deep neural policy in RoadRunner. Left: Adversarially trained. Right: Vanilla trained.}
    \label{roadhmap}
\end{figure*}

\begin{figure*}[h!]
\footnotesize
\begin{center}
\stackunder[6pt]{\includegraphics[scale=0.34]{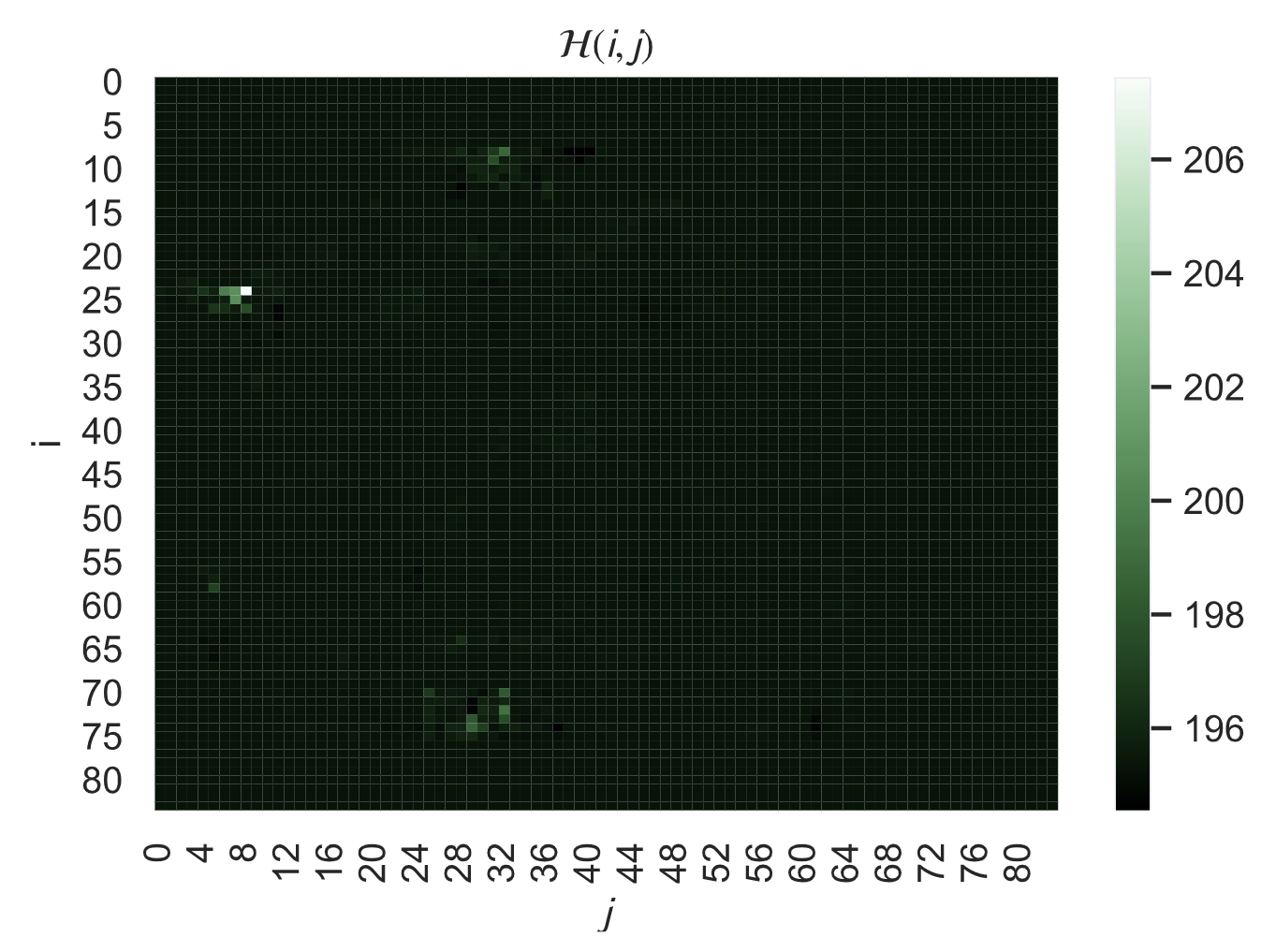}}{Adversarially Trained}
\hskip 0.1pt
\stackunder[6pt]{\includegraphics[scale=0.33]{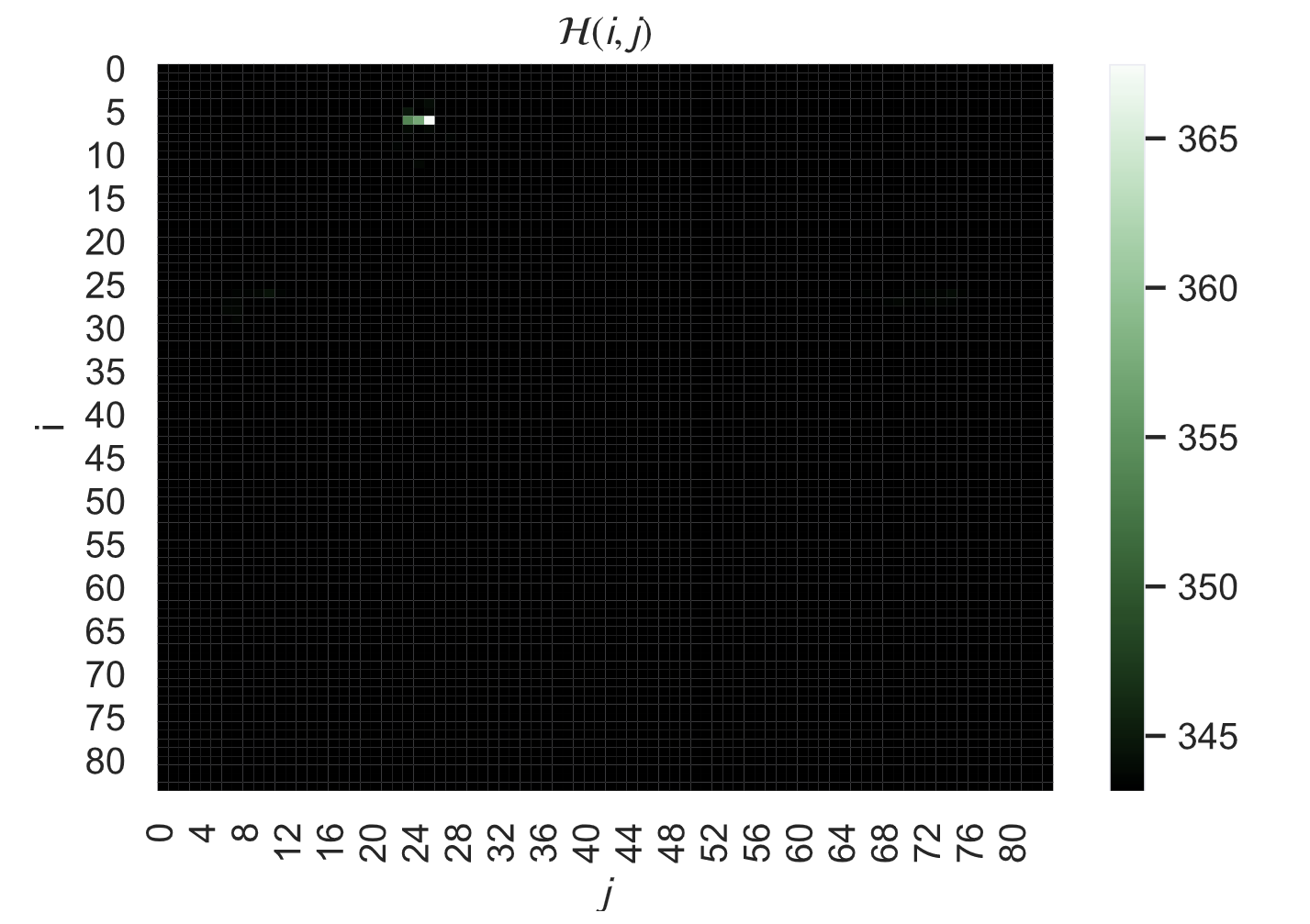}}{Vanilla Trained}
\end{center}
  \caption{HMAP $\mathcal{H}(i,j)$ heatmaps for state-of-the-art adversarially trained deep neural policy and vanilla trained deep neural policy in BankHeist. Left: Adversarially trained. Right: Vanilla trained.}
    \label{bankhmap}
\end{figure*}

\begin{figure*}[h!]
\begin{multicols}{2}
\includegraphics[width=1.1\linewidth]{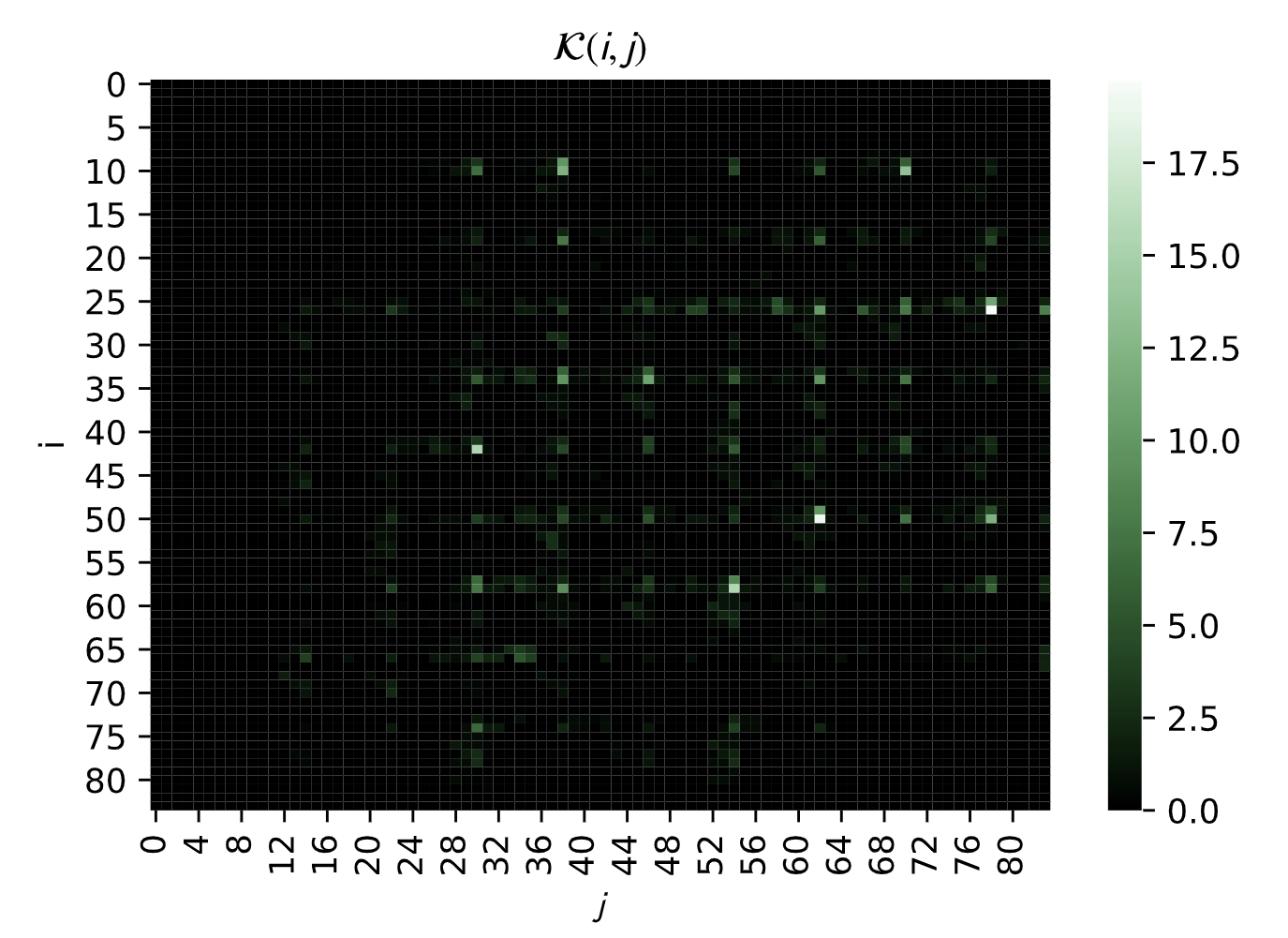}\par
\includegraphics[width=1.1\linewidth]{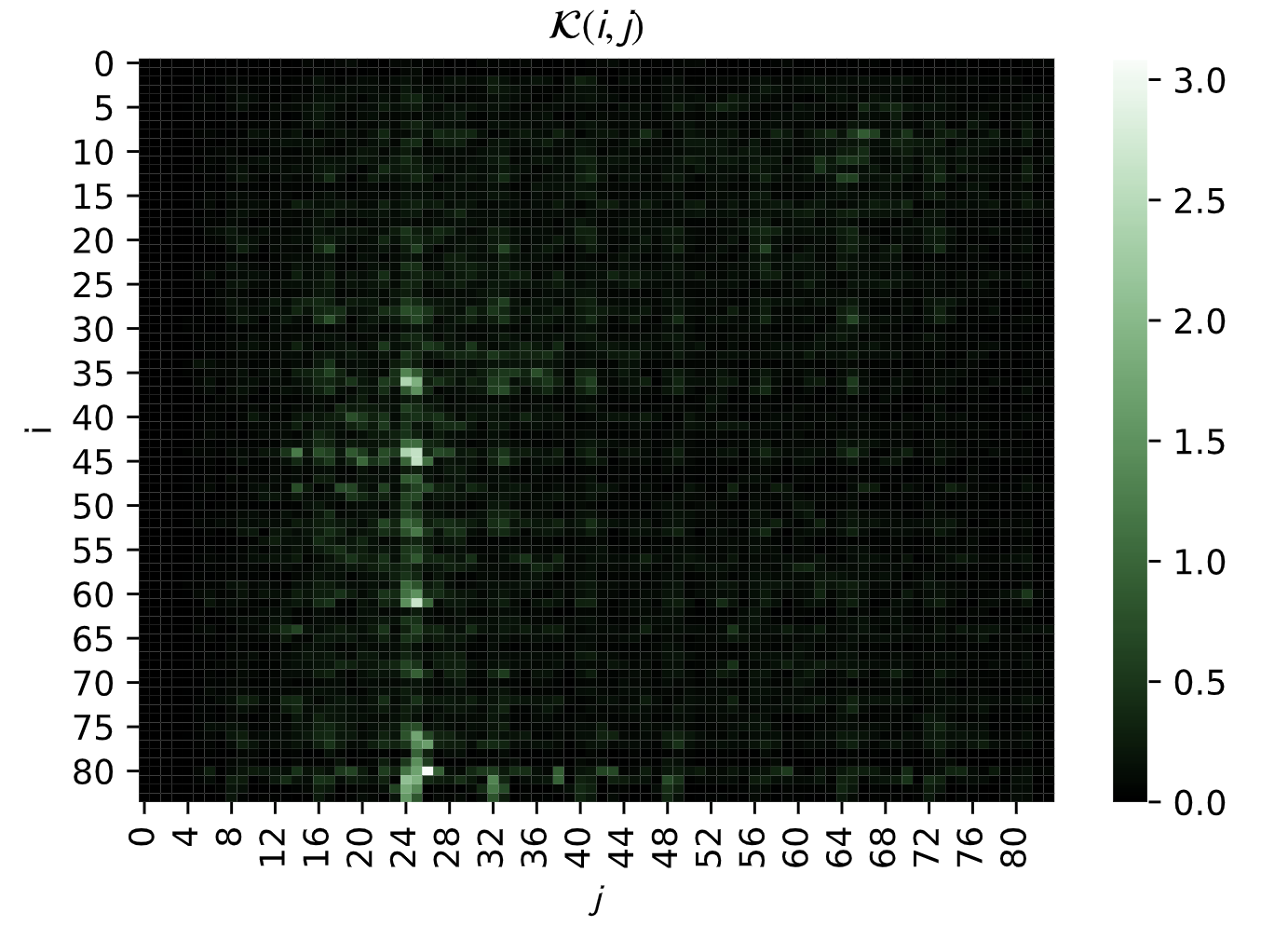}\par
\end{multicols}
\vspace{-0.4cm}
\caption{KMAP $\mathcal{K}(i,j)$ heatmaps for state-of-the-art adversarially trained deep neural policy and vanilla trained deep neural policy in Freeway. Left: Adversarially trained (SA-DDQN). Right: Vanilla trained (DDQN).}
\label{freekmap}
\end{figure*}

\begin{figure*}[h!]
\begin{multicols}{2}
\includegraphics[width=1.1\linewidth]{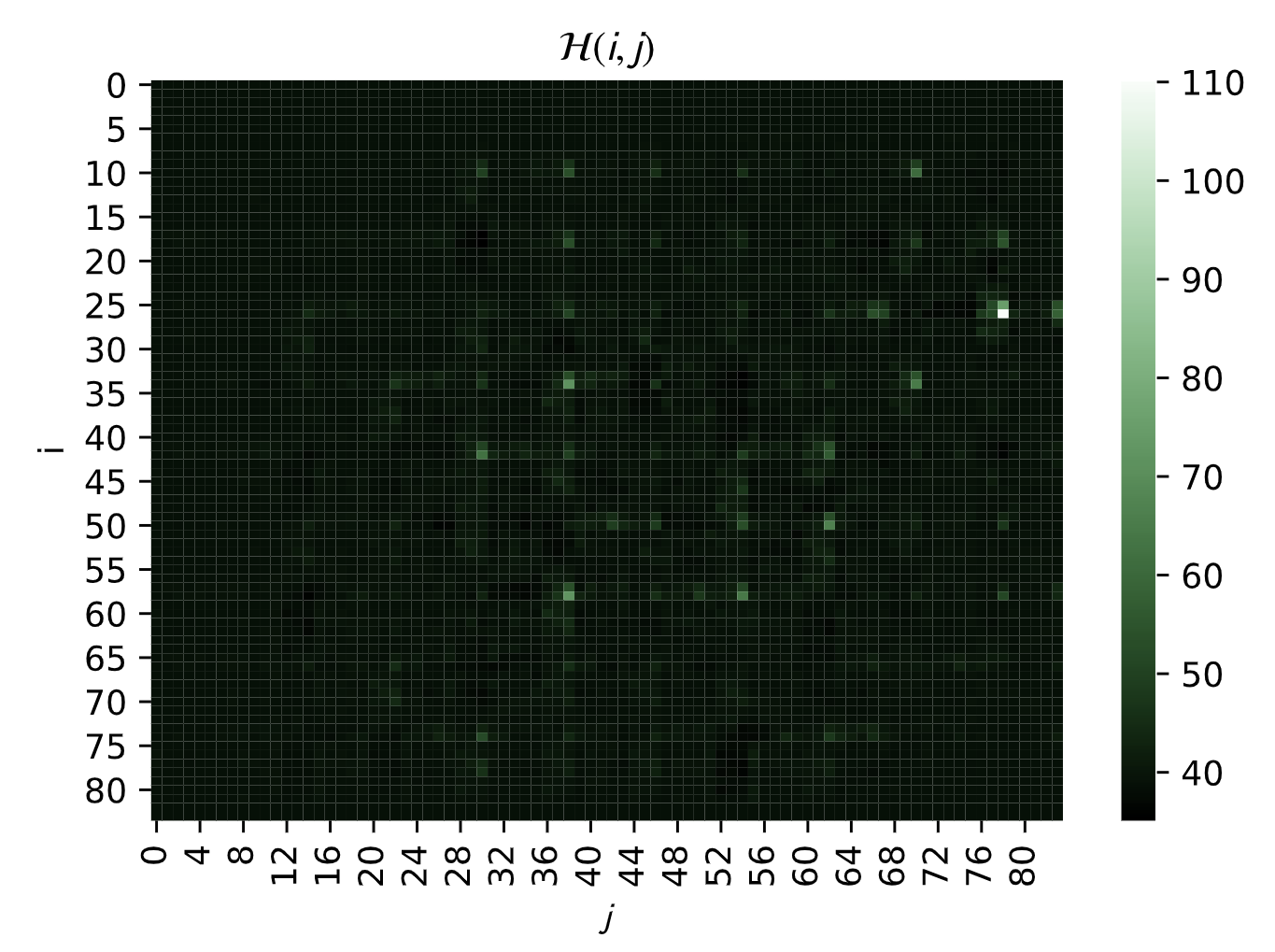}\par
\includegraphics[width=1.1\linewidth]{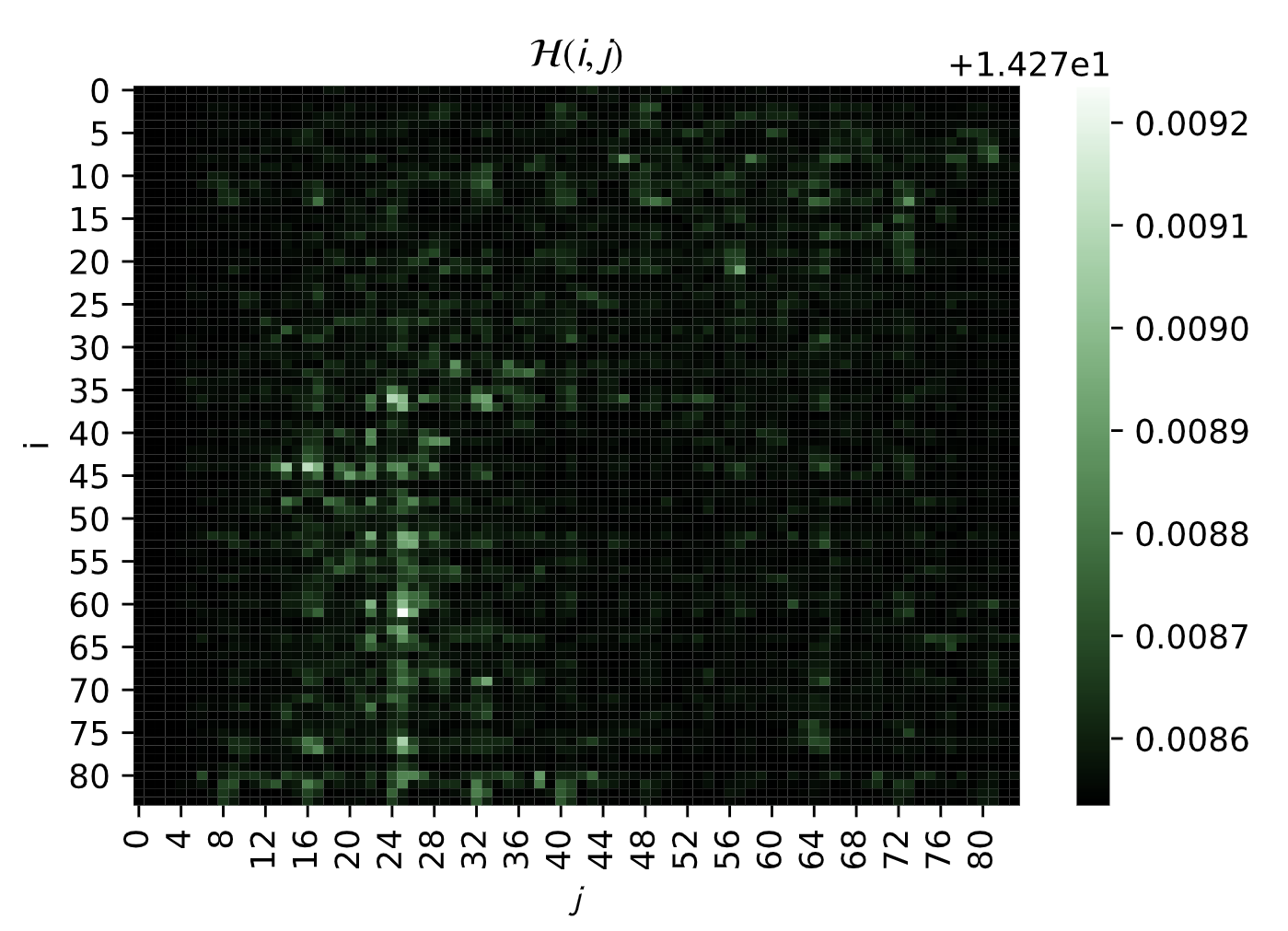}\par
\end{multicols}
\vspace{-0.4cm}
\caption{HMAP $\mathcal{H}(i,j)$ heatmaps for state-of-the-art adversarially trained deep neural policy and vanilla trained deep neural policy in Freeway. Left: Adversarially trained (SA-DDQN). Right: Vanilla trained (DDQN).}
\label{freehmap}
\end{figure*}
in this setting we observe that the vulnerabilities towards a certain set of features remians the same with adversarial training.

Figure \ref{freehmap} and Figure \ref{freekmap} show heatmaps of KMAP $\mathcal{K}(i,j)$ and HMAP $\mathcal{H}(i,j)$ for Freeway. We observe that while the KMAP $\mathcal{K}(i,j)$ pattern for the vanilla trained agent lies on the portion of the input where the optimal policy is executed by the agent, the KMAP $\mathcal{K}(i,j)$ for the adversarially trained deep neural policy has a straightforward grid pattern. Based on these results, we hypothesize that adversarial training decouples vulnerability from the features relevant to the optimal policy learned by the agent.

The decoupling of relevant features and vulnerability can be seen as an additional way in which adversarial training shifts the vulnerabilities of deep neural policies. This complements the results of Section \ref{fourier}, where we observe a vulnerability shift by looking at worst-case $\ell_p$-norm bounded perturbations, and observing that these perturbations are more concentrated on lower frequencies in adversarially trained agents.

While visual observation indicates very different vulnerability patterns for these two disjoint training strategies, we also introduce a quantitative metric to compare the results of KMAP and HMAP for vanilla and adversarially trained agents. In particular,  we use the ratio of the $\ell_1$ and $\ell_2$ norms to measure the sparsity via,
\begin{equation}
S(\mathcal{K}) = \dfrac{\norm{\mathcal{K}}_1}{\norm{\mathcal{K}}_2}.
\end{equation}
Here smaller values of $S(\mathcal{K})$ correspond to sparser vulnerability patterns.
We also measure how spread out the vulnerability pattern is via the entropy of the softmax of $\mathcal{K}(i,j)$,
\begin{table*}[h!]
\caption{Sparsity results of KMAP $\mathcal{K}(i,j)$ and HMAP $\mathcal{H}(i,j)$ for adversarially trained and vanilla trained deep neural policies.}
% \vskip 0.04in
\label{spar}
\centering
\scalebox{0.95}{
\begin{tabular}{lccccr}
\toprule
Training Method &  Vanilla Trained
				  &  Adversarially Trained
				  &  Vanilla Trained
				  &  Adversarially Trained \\
Sparsity    &   $S(\mathcal{K})$
				  &  $S(\mathcal{K})$
				  &  $S(\mathcal{H})$
				  &   $S(\mathcal{H})$\\
\midrule
Freeway     & 53.7272  & 20.4641   &   83.9999        &83.91587 \\
BankHeist   & 38.4085   & 4.8812    &   83.99999      &83.99994    \\
RoadRunner& 33.1493   & 11.3216  &   83.999992     & 83.999993   \\
Pong 			&  49.7993   & 1.9508   &   83.99999       & 83.9999  \\
\bottomrule
\end{tabular}
}
\end{table*}
\begin{equation}
H(\mathcal{K}) = -\sum_{i,j}\softmax(\mathcal{K})_{i,j}\log(\softmax(\mathcal{K})_{i,j})
\end{equation}
In Table \ref{spar} and Table \ref{ent} we show the sparsity and entropy results respectively for KMAP $\mathcal{K}(i,j)$ and HMAP $\mathcal{H}(i,j)$ for adversarially trained deep neural policies and vanilla trained deep neural policies. We observe that for KMAP the vulnerability of adversarially trained models with respect to features are more sparse than the vanilla trained agents. The

\begin{table*}[h!]
\caption{Entropy of KMAP $\mathcal{K}(i,j)$ and HMAP $\mathcal{H}(i,j)$ for adversarially trained deep neural policies and vanilla trained deep neural policies.}
% \vskip 0.04in
\label{ent}
\centering
\scalebox{0.96}{
\begin{tabular}{lccccr}
\toprule
Training Method &  Vanilla Trained
				  &  Adversarially Trained
				  &  Vanilla Trained
				  &  Adversarially Trained \\
Entropy    &   $H(\mathcal{K})$
				  &  $H(\mathcal{K})$
				  &  $H(\mathcal{H})$
				  &   $H(\mathcal{H})$\\
\midrule
Freeway     & 8.8287  & 0.0807 & 8.8616 &0.5677 \\
BankHeist   & 0.0055  & 1.542$e^{-20}$   & 8.8616 & 0.54405 \\
RoadRunner&1.2346   & 0.6973  &   8.8610   & 8.8608  \\
Pong 			&8.8615   & 8.5658  &   8.8616   & 8.8615 \\
\bottomrule
\end{tabular}
}
\end{table*}

results for HMAP are more mixed, and it is often barely possible to detect the sparsity difference via $S(\mathcal{H})$ and only possible in half of the games via $H(\mathcal{H})$. In general, KMAP $\mathcal{K}(i,j)$ provides a better estimation of sensitivity of deep neural policies to individual pixel changes than HMAP $\mathcal{H}(i,j)$. While KMAP $\mathcal{K}(i,j)$ captures the actual impact of the feature change on the decision of the deep neural policy HMAP $\mathcal{H}(i,j)$ captures the difference between the softmax policy distributions $\pi(s,a)$ and $\pi(Z_{i,j}(s),a)$, which do not necessarily correspond to the decisions made by the neural policy.

\section{Conclusion}

In this paper we focused on investigating the vulnerabilities of deep neural policies with respect to their inputs. We examined the vulnerability shifts between state-of-the-art adversarially trained deep neural policies and vanilla trained policies. First, we investigate through worst-case $\ell_p$-norm bounded distributional shift. We explored and compared the frequency domain of the perturbations computed from state-of-the-art adversarially trained neural policies and vanilla trained neural policies. We found that the perturbations computed from adversarially trained models were more concentrated in lower frequencies compared to the vanilla trained neural policies. Second, we propose two different algorithms that we call KMAP and HMAP to detect vulnerabilities with respect to input in deep neural policies. We compare the state-of-the-art adversarially trained neural policies and vanilla trained neural policies with our proposed methods KMAP and HMAP via several experiments in various environments. We found that while adversarial training removes sensitivity to certain features, it builds sensitivity towards a new set of features. We believe this work lays out the vulnerabilities of adversarially trained neural policies in a systematic way, and can be an initial step towards building robust and reliable deep reinforcement learning agents.

\bibliography{example_paper}

\end{document}